\documentclass[10pt,conference]{IEEEtran}

\usepackage{amsmath}
\usepackage{balance}
\usepackage{booktabs}
\usepackage{cite}
\usepackage{dblfloatfix}
\usepackage{graphicx}
\usepackage{hyperref}
\usepackage{multicol}
\usepackage{multirow}
\usepackage[position=bottom]{subfig}
\usepackage{wrapfig}
\usepackage[dvipsnames]{xcolor}

\usepackage{cleveref}

\def\BibTeX{{\rm B\kern-.05em{\sc i\kern-.025em b}\kern-.08em
    T\kern-.1667em\lower.7ex\hbox{E}\kern-.125emX}}

\graphicspath{{fig/}}

\newcommand{\name}{{Taxon}}

\Crefname{equation}{Eq.}{Eqs.}
\Crefname{figure}{Fig.}{Figs.}
\Crefname{section}{Sec.}{Secs.}
\Crefname{table}{Tab.}{Tabs.}

\IEEEoverridecommandlockouts

\begin{document}
\title{\name{}: Hierarchical Tax Code Prediction \\ with Semantically Aligned LLM Expert Guidance}


\author{
  \IEEEauthorblockN{
    Jihang Li\textsuperscript{1,$\dagger$}\thanks{$\dagger$ Equal contribution.},
    Qing Liu\textsuperscript{2,$\dagger$},
    Zulong Chen\textsuperscript{2,*}\thanks{* Corresponding authors.},
    Jing Wang\textsuperscript{2},
    Wei Wang\textsuperscript{2},
    Chuanfei Xu\textsuperscript{3,*},
    Zeyi Wen\textsuperscript{1,*}
  }

  \IEEEauthorblockA{
    \textsuperscript{1}\textit{The Hong Kong University of Science and Technology (Guangzhou), Guangzhou, China} \\
    \textsuperscript{2}\textit{Alibaba Group, Hangzhou, China} \\
    \textsuperscript{3}\textit{Guangdong Laboratory of Artificial Intelligence and Digital Economy (Shenzhen), Shenzhen, China} \\
    zulong.czl@alibaba-inc.com, xuchuanfei@gml.ac.cn, wenzeyi@hkust-gz.edu.cn
  }
}

\maketitle



\begin{abstract}
  Tax code prediction is a crucial yet underexplored task in automating
  invoicing and compliance management for large-scale e-commerce platforms.
  Each product must be accurately mapped to a node within a multi-level
  taxonomic hierarchy defined by national standards, where errors lead to
  financial inconsistencies and regulatory risks. This paper presents \name{},
  a semantically aligned and expert-guided framework for hierarchical tax code
  prediction. Taxon integrates (i) a feature-gating mixture-of-experts
  architecture that adaptively routes multi-modal features across taxonomy
  levels, and (ii) a semantic consistency model distilled from large language
  models acting as domain experts to verify alignment between product titles
  and official tax definitions. To address noisy supervision in real business
  records, we design a multi-source training pipeline that combines curated tax
  databases, invoice validation logs, and merchant registration data to provide
  both structural and semantic supervision. Extensive experiments on the
  proprietary TaxCode dataset and public benchmarks demonstrate that
  \name{} achieves state-of-the-art performance, outperforming strong
  baselines. Further, an additional full hierarchical paths reconstruction
  procedure significantly improves structural consistency, yielding the highest
  overall $F_{1}$ scores. \name{} has been deployed in production within
  Alibaba's tax service system, handling an average of about five million tax code
  queries per day and reaching peak volumes up to fifteen million requests during
  business event with improved accuracy, interpretability, and robustness.
\end{abstract}

\section{Introduction}
\label{sec:introduction}
Modern e-commerce platforms in China process millions of transactions daily,
each requiring the correct assignment of a tax code for invoicing and
regulatory compliance. This task, known as \textit{hierarchical tax code
prediction}, maps a product--represented by metadata such as title, category,
and attributes--to a specific code within a tree taxonomy defined by the State
Taxation Administration. With up to ten hierarchical levels and over four
thousand leaf categories, manual classification is both costly and error-prone,
while errors directly cause financial discrepancies or compliance violations.

\subsection{Background}
\label{sec:background}
\begin{figure}[t]
  \centering
  \includegraphics[width=\columnwidth]{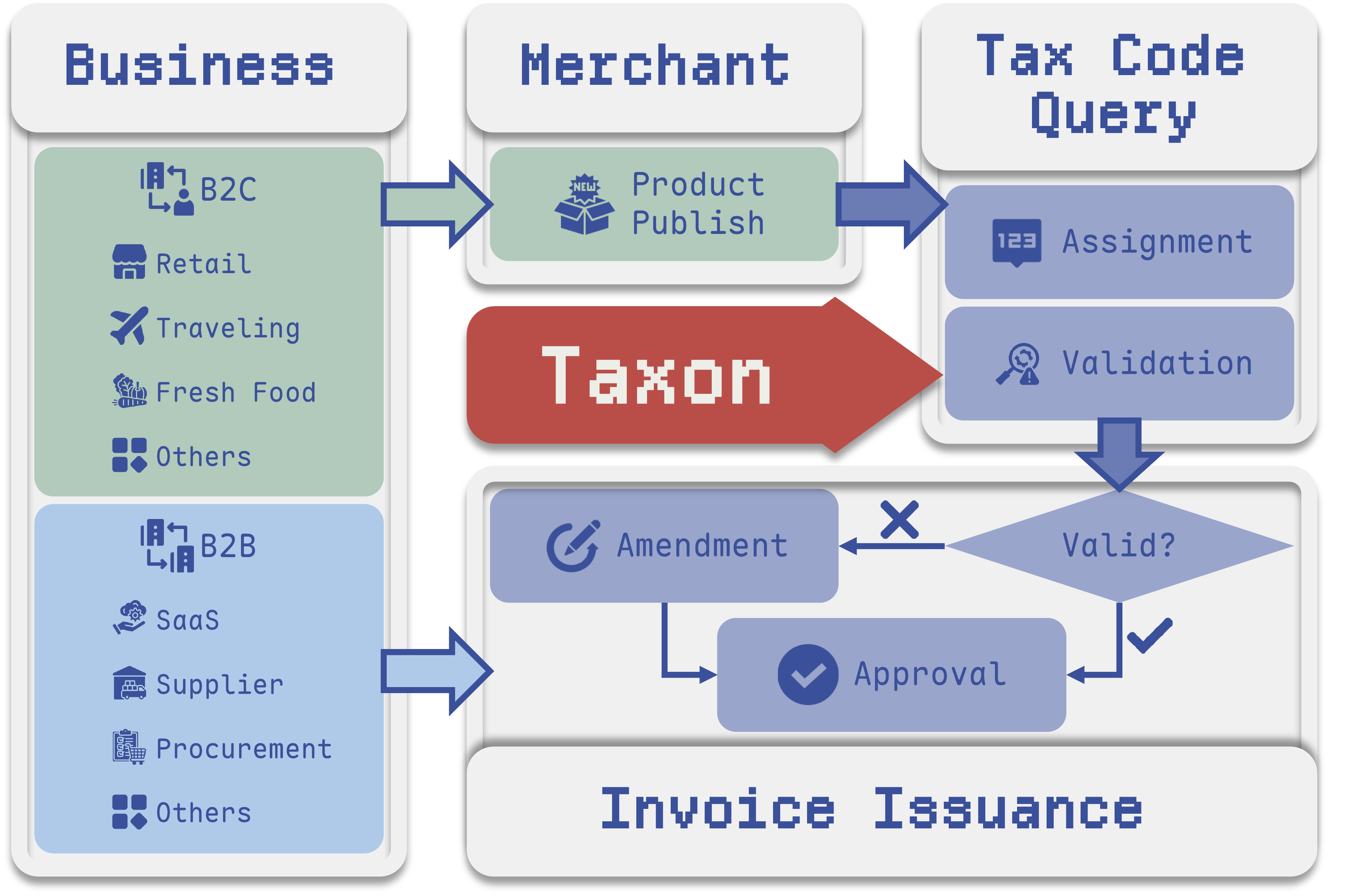}

  \caption{An illustration of tax code prediction in e-commerce.}
  \label{fig:business}
\end{figure}

Tax code prediction is embedded within a complex commercial and regulatory
ecosystem that connects merchants, consumers, and tax authorities. As shown in
\Cref{fig:business}, this ecosystem spans two major operational scenarios:
business-to-customer (B2C) and business-to-business (B2B), together covering
the end-to-end workflow on large e-commerce platforms. In both scenarios, each
product or transaction follows a standardized pipeline of product publication,
tax code query, validation, and invoice issuance. These business flows motivate
our design by revealing where automation can most effectively reduce manual
workload and improve consistency. The following subsections detail the B2C and
B2B scenarios.


\paragraph{B2C Scenario}
The B2C operations encompass a variety of online services that support
customers' daily needs, ranging from retail (e.g., Tmall) and traveling (e.g.,
Fliggy) to fresh food (e.g., Freshippo), digital content (e.g., Youku), and
local services (e.g., Ele.me). In these settings, tax code assignment occurs
during the \textit{product publish} phase, when third-party merchants register
new items on the platform. Each item must be associated with a valid tax code
before it can go online. To assist merchants, traditionally, the platform
performs a three-stage progressive query against the internal tax code database
organized as key-value pairs:  (i) query by \textit{product title} (e.g.,
``Fully Automatic Washing Machine 3 L''); (ii) if no match is found, fall back
to \textit{product category} (e.g., ``Home Appliances''); and (iii) if still
unmatched, a tax code must be selected manually from a list provided in the
merchant console. Once a code is identified at any stage, it is returned to the
merchant for confirmation and recorded in the listing.

\paragraph{B2B Scenario}
In contrast, the B2B branch of the ecosystem handles internal transactions. The
platform itself serves as a buyer or service provider for scenarios such as
supplier management, asset procurement, and enterprise SaaS. Here, tax code
assignment is performed internally rather than by external merchants. When a
business transaction occurs, the system proceeds to the \textit{invoice
validation} process, mirroring the same three-step query pipeline used in B2C.
If the retrieved code is valid and consistent with the record, the transaction
proceeds automatically to \textit{invoice issuance} and approval. Otherwise,
any absent or inconsistent code triggers an \textit{amendment} step handled by
in-house tax experts before final approval. This closed-loop workflow, as shown
in \Cref{fig:business}, ensures that both B2C and B2B channels maintain
traceable and auditable tax code assignments with minimal human intervention
and compliant with official regulations.

\subsection{Motivation}
\label{sec:motivation}
However, maintaining and validating tax codes across large business
operations remains challenging and error-prone. The tax code database is
curated manually and periodically updated whenever new products or categories
emerge. This process is labor-intensive and often inconsistent, especially when
product metadata is incomplete or ambiguous. Such inconsistencies propagate to
erroneous code assignments, leading to financial discrepancies and compliance
risks. As a result, reliance on human experts limits efficiency, accuracy, and
scalability in high-volume e-commerce operations.

These issues motivate the need for an automated hierarchical tax code
prediction system that can assist merchants and internal auditors in assigning
codes accurately and consistently from product metadata. However, automatic
prediction introduces additional challenges stemming from the hierarchical and
heterogeneous nature of real-world data:

\begin{itemize}
  \item \textit{High-dimensional, incomplete features.} The model must
    integrate textual and structured modalities--titles, categories, prices,
    and organizational attributes--despite missing or unreliable fields that
    violate the clean-text assumption of traditional classifiers.
  \item \textit{Deep and large label hierarchy.} The official taxonomy spans up
    to ten levels and over four thousand leaf nodes. Flat classifiers ignoring
    parent-child constraints accumulate cascading errors. Effective models must
    exploit hierarchical dependencies to prune irrelevant subtrees and maintain
    valid paths.
  \item \textit{Semantic ambiguity and label relationships.} Many tax codes
    have overlapping meanings or “other” categories that differ only by
    fine-grained material or functional nuances. Capturing sibling and
    parent-child relations is essential to avoid inconsistent predictions.
  \item \textit{Noisy and inconsistent supervision.} Historical records contain
    human and procedural inconsistencies, and long-tail imbalance amplifies
    noise. The model must remain robust through confidence calibration,
    hierarchy-aware losses, and semantic validation.
\end{itemize}

\subsection{Proposal}
\label{sec:proposal}
Tax code prediction faces the same hierarchical and semantic complexities that
make manual maintenance difficult. It can be formulated as a problem of
\textit{hierarchical text classification} (HTC), where each product is mapped
to a node in a multi-level taxonomy defined by the \textit{Goods and Services
Tax Classification Catalogue}~\cite{sta2017goods}. Like the Harmonized System
(HS) used in customs and trade, it demands learning fine-grained category
boundaries that respect strict parent-child dependencies and semantic
constraints.

Earlier works on tax or HS code prediction relied on conventional text
classifiers~\cite{luppes2019classifying,shubham2023ensemble,liao2024enhanced}.
Although they automated parts of the process, these models remained fragile in
practice: errors accumulated along deeper hierarchy levels, semantic
inconsistencies with official definitions persisted under noisy supervision,
and their interpretability for business auditing was limited. They also
struggled to adapt to heterogeneous data sources, hindering scalability in
enterprise systems.

To address these limitations, we propose \textbf{\name{}} for hierarchical tax
code prediction. \name{} integrates structural reasoning, semantic alignment,
and data robustness through three complementary components:

\begin{itemize}
  \item \textbf{Hierarchical feature-gating mixture-of-experts (MoE)}
    architecture that captures multi-level dependencies and adaptively routes
    features across taxonomy levels to maintain logical path consistency;
  \item \textbf{Semantic consistency-assisted prediction} guided by large
    language models (LLMs) acting as domain experts, which distill semantic
    alignment between product titles and official tax definitions through
    expert judgments;
  \item \textbf{Multi-source training pipeline} integrating over eight million
    business records to provide diverse and reliable supervision, effectively
    mitigating noise and handling incomplete product metadata.
\end{itemize}

In contrast to traditional HTC approaches, \name{} explicitly bridges taxonomic
structure, semantic meaning, and real-world business workflows, forming a
unified architecture that is both interpretable and robust. Deployed within
Alibaba's enterprise tax service system, \name{} now processes an average of about five
million tax code queries per day across multiple business units, achieving high
accuracy, scalability, and transparency under real operational conditions.

Our preliminary analyses further reveal that the primary source of
classification errors lies not in misidentifying the correct leaf category but
in violating the hierarchical structure that connects intermediate nodes. In
other words, residual errors mainly arise from \textit{structural
inconsistency} along the predicted path rather than \textit{semantic
misunderstanding} at the leaf level. This observation highlights a common
limitation of existing HTC methods, which often entangle structural and
semantic errors within a single objective. Motivated by this, we explicitly
decouple the two aspects: (i) maximizing semantic precision at the leaf level,
and (ii) designing an independent mechanism, ``\textbf{RePath}''
(\underline{re}constructing \underline{path} from leaf), to ensure structural
consistency of the predicted hierarchy.

\subsection{Contribution and Organization}
\label{sec:contribution-organization}
This work makes the following key contributions:

\begin{enumerate}
  \item We formalize the end-to-end workflow of hierarchical tax code
    prediction across both B2C and B2B operations, clarifying how automated
    inference integrates with expert validation to ensure compliance and
    consistency in real business processes.
  \item We present \name{}, a semantically aligned and expert-guided
    hierarchical prediction framework that unifies hierarchical feature-gating
    and LLM-assisted semantic supervision within a single, end-to-end system.
  \item We design an LLM-distilled semantic labeling pipeline that bridges
    structured tax definitions and unstructured product text, enabling robust
    supervision even under noisy or incomplete annotations.
  \item We demonstrate large-scale industrial deployment and comprehensive
    evaluation across internal and public benchmarks, showing consistent gains
    over strong hierarchical and semantic baselines and validating the system's
    reliability in daily production environments.
\end{enumerate}

The remainder of this paper is organized as follows: \Cref{sec:methodology}
presents our framework and training pipeline; \Cref{sec:experiments} reports
the experimental evaluation; \Cref{sec:related} reviews related work; and
\Cref{sec:conclusion} concludes our work.


\section{Methodology}
\label{sec:methodology}
Hierarchical tax code prediction maps a product to a node in a multi-level
taxonomy, requiring both structural validity and semantic correctness. We
address this with a framework that couples a hierarchical feature-gating MoE
for multi-level prediction with LLM-guided semantic consistency supervision to
align product descriptions with official tax definitions.


\subsection{Framework Overview}
\label{sec:framework-overview}
We formulate tax code prediction as a hierarchical multi-feature classification
problem that leverages product information and structured business metadata.
The framework comprises (i) a text encoder that captures the semantic features
of product titles and categories, and (ii) a hierarchical feature-gating
mixture-of-experts (MoE) classifier that predicts the most appropriate tax code
at each level of the taxonomy.

\begin{figure}[t]
  \centering
  \includegraphics[width=\columnwidth]{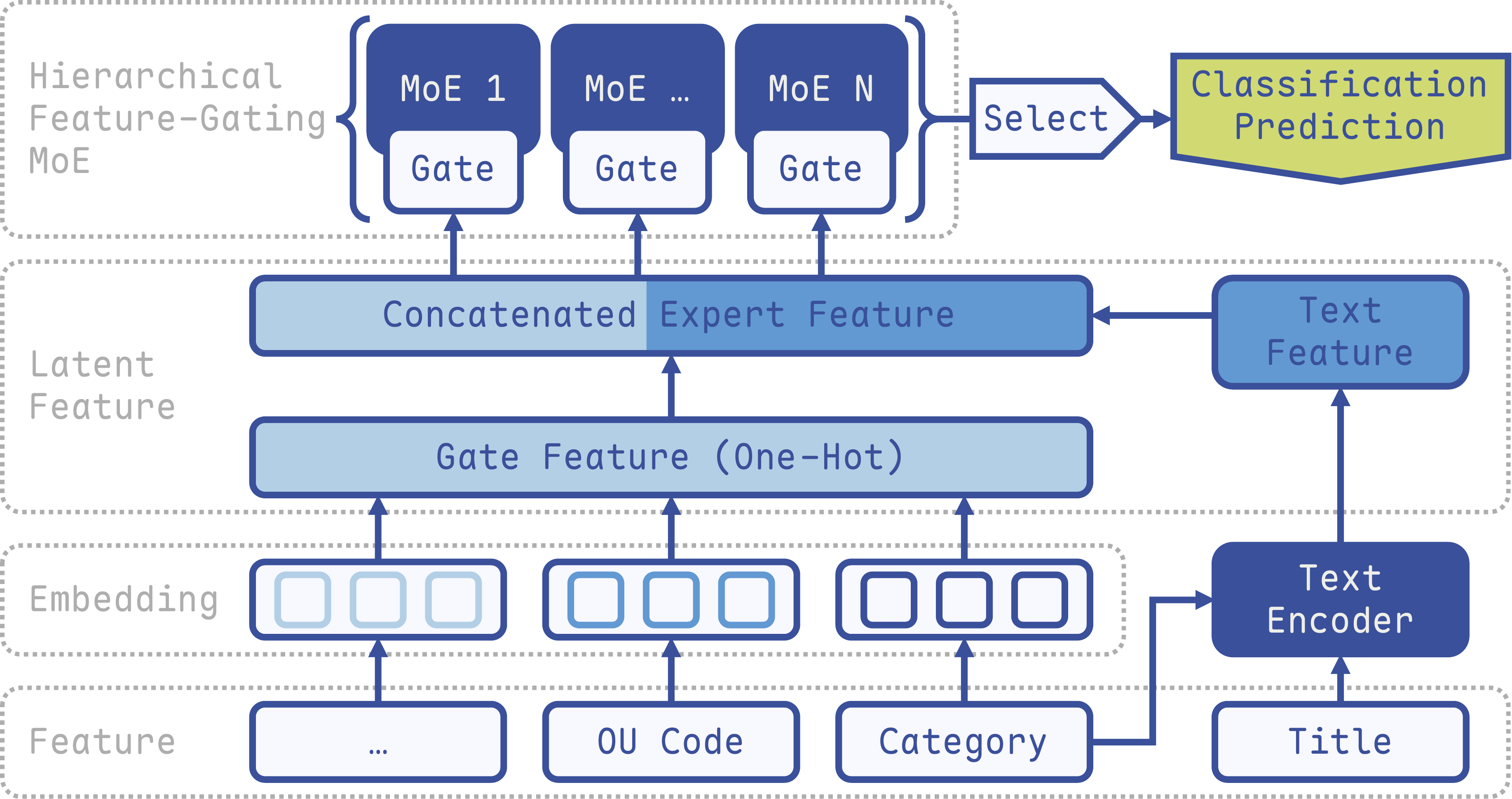}

  \caption{Overview of the proposed framework. The system processes both
  textual and structured features through a hierarchical MoE to predict tax
  codes at multiple levels.}
  \label{fig:model-inference}
\end{figure}

As shown in \Cref{fig:model-inference}, the system processes the input features
in a bottom-up manner, where product titles, categories, and other relevant
attributes are first encoded into latent representations. These features are
then routed to specialized experts through a gating mechanism, which adaptively
assigns them to the most relevant experts based on feature semantics. The final
tax code is determined by aggregating predictions from the experts at various
levels of the hierarchy.

During inference, the system prioritizes leaf-level predictions with the
highest confidence. If no valid leaf-level prediction is found, the model
selects the deepest node along the predicted path with the highest confidence
score. This approach ensures that the output respects the hierarchical
structure of the tax code while maintaining robustness across varying levels of
granularity.

\paragraph{Input Features}
The framework takes as input both unstructured and structured product
information. The primary textual inputs are the product title and category
name, which describe the product's key attributes. These are complemented by
structured business metadata, including the business unit (BU) code,
organization unit (OU) code, and system code (e.g., ``TD'', ``HP6'', and
``TM\_3C''). Although these structured identifiers carry little inherent
semantics, they provide useful contextual cues that distinguish product
distributions across business units. Additional features such as category
property values (CPVs)--for example, ``hard disk type'' for computers or
``material'' for furniture--can also be incorporated when available to enrich
the input space.

\paragraph{Feature Encoding}
The textual features (\textit{title} and \textit{category}) are processed
through a pre-trained text encoder such as BERT~\cite{devlin2018bert},
TextCNN~\cite{kim2014conv}, or XLNet~\cite{yang2019xlnet}, which captures
contextual semantics in a latent representation. Structured features are mapped
to learnable embeddings and represented as one-hot vectors to preserve their
categorical distinctions. The encoded text representation and structured
embeddings are concatenated into a unified feature vector, forming a
comprehensive representation of each product that combines semantic meaning
with business context.

\paragraph{Hierarchical Feature-Gating MoE}
The concatenated feature vector serves as the input to a hierarchy of
feature-gating MoE modules, where each module corresponds to one level in the
tax code taxonomy. Given that the official taxonomy contains ten levels, we
deploy ten MoE modules in parallel, each contains eight experts and is
responsible for predicting labels within its respective level. Each expert
specializes in a subset of product types or hierarchical levels, enabling the
overall model to capture level-specific dependencies while maintaining
scalability across a large label space. The gating network of each MoE is
trained to generate routing weights from the structured one-hot features and
produces a softmax distribution over experts, guiding the input representation
toward the most relevant ones. All expert and gate parameters are initialized
with Xavier initialization, and optimized end-to-end via backpropagation.

\paragraph{Final Prediction}
Since the hierarchical MoE produces multiple predictions across taxonomy
levels, a selection strategy is applied to determine the final output. The
framework first prioritizes leaf-level predictions with the highest confidence,
as leaf nodes represent the most specific tax codes. If no confident leaf
prediction exists, the model selects the deepest valid node along the predicted
path with the highest probability score. This strategy ensures that the final
output adheres to the hierarchical taxonomy, maintaining logical path
consistency while preserving robustness against uncertain or incomplete
predictions.

\subsection{Hierarchical-Semantic Training}
\label{sec:model-training}
\begin{figure}[!htb]
  \centering
  \includegraphics[width=\columnwidth]{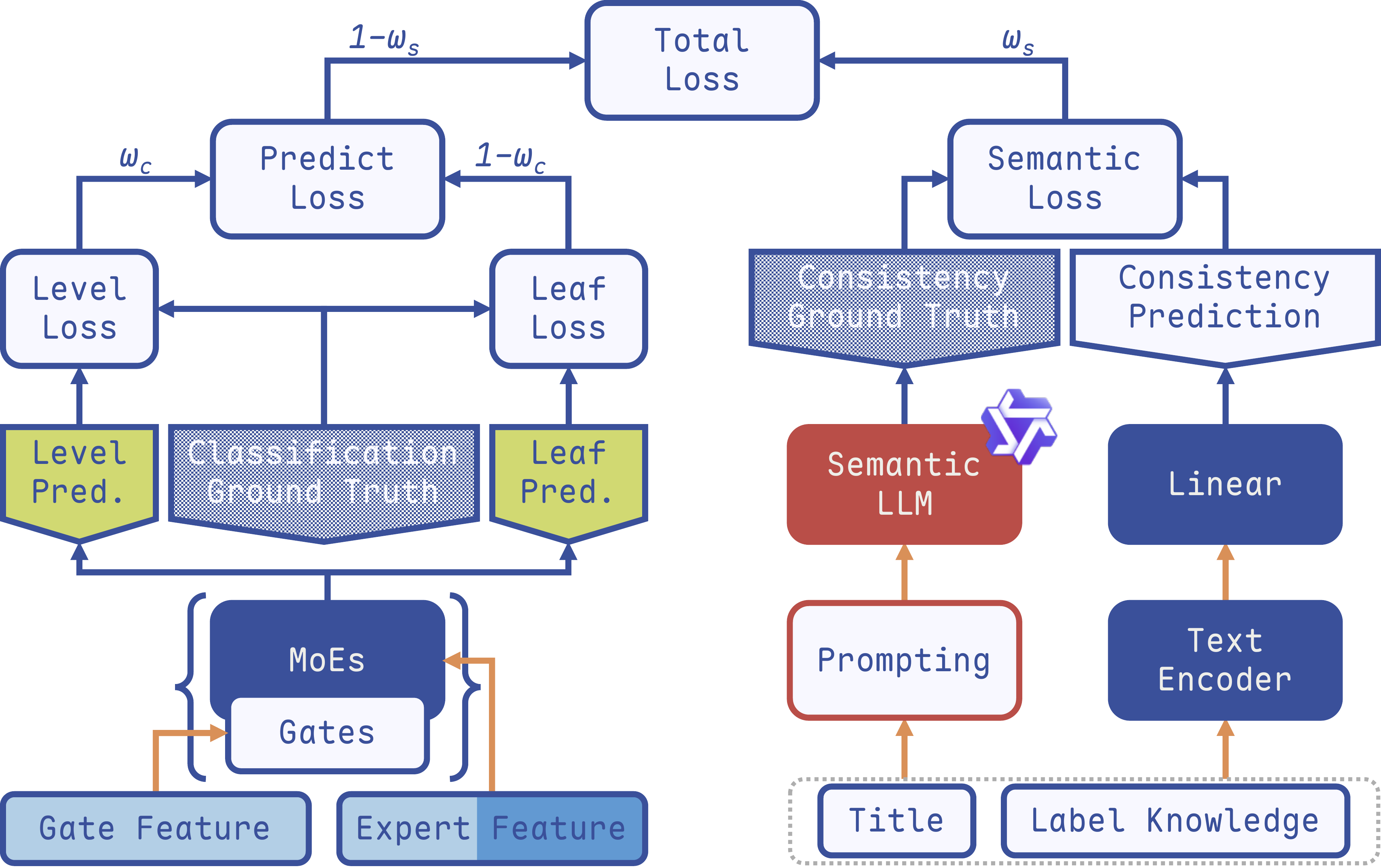}

  \caption{Overview of the training workflow, integrating hierarchical
  feature-gating MoE with LLM-assisted semantic consistency. The multi-stage
  pipeline ensures robust learning across the tax code hierarchy.}
  \label{fig:model-training}
\end{figure}

Our framework adopts an training paradigm that optimizes two complementary
objectives: \textit{hierarchical classification} and \textit{semantic
consistency}. As illustrated in \Cref{fig:model-training}, it couples a
hierarchical feature-gating MoE with an LLM-guided semantic alignment branch in
a multi-task pipeline. The hierarchical component captures dependencies across
taxonomy levels to preserve structural integrity, while the semantic branch
enforces alignment between product descriptions and tax definitions through
expert-derived signals. In this process, a large language model acts as a
\textit{distiller of domain knowledge}, translating its reasoning into
lightweight, task-specific supervision. By judging whether each product title
matches its tax code definition, the LLM provides domain-grounded feedback
distilled into a compact \textit{Semantic LLM}, enabling efficient and
continual learning of hierarchical-semantic consistency.


On the left side of \Cref{fig:model-training}, the hierarchical feature-gating
MoE models dependencies across taxonomy levels. During training, the gating
network learns routing weights from one-hot business metadata (e.g., product
category and system codes), directing each input to the most relevant experts.
Each expert specializes in a subset of levels and produces predictions for its
assigned scope. The model minimizes a hierarchical classification loss across
all levels to preserve the overall structural consistency of the tax code
hierarchy.

On the right side of \Cref{fig:model-training}, the semantic alignment branch
enforces consistency between product titles and official tax code definitions.
A title and its assigned tax code are fed into a semantic encoder: one branch
encodes the definition into a reference embedding, while the other projects the
product representation into a predicted embedding. Their divergence defines the
semantic consistency loss, guiding towards taxonomically valid and semantically
coherent predictions.

The total loss combines the hierarchical classification and semantic
consistency terms, ensuring that the model jointly captures structural accuracy
and semantic validity. Training proceeds through backpropagation with tunable
weights balancing the two objectives. The following subsection details the loss
formulation, dataset construction, and training pipeline, including the
semantic LLM distillation process that transfers expert-level judgments into
lightweight supervision for large-scale learning.

\subsubsection{Hierarchical Classification Loss}
Given an input $x$ with ground-truth $N$-level labels $\mathbf{y} = \{y_{1},
y_{2}, \dots, y_{N}\}$, the framework produces predictions ${\hat{y}}_{i}$ at
each level $i \in \{1, 2, \dots, N\}$ of the tax code hierarchy ($N = 10$ in
our case). Each level is supervised with a cross-entropy
loss~\cite{rubinstein2004cross}:
\begin{equation}
  \mathcal{L}_{i} = \mathrm{cross\_entropy}(\hat{y}_{i}, y_{i}).
  \label{eqn:level-loss}
\end{equation}
Among the $N$ levels, one corresponds to the leaf node, which provides the most
fine-grained supervision. The overall hierarchical classification loss is
computed as a weighted sum of all intermediate-level losses and the leaf-level
loss:
\begin{equation}
  \mathcal{L}_{c} = \omega_{c} \sum^{N - 1} \mathcal{L}_{i} + (1 - \omega_{c}) \mathcal{L}_{\mathrm{leaf}},
  \label{eqn:classification-loss}
\end{equation}
where the weight $\omega_{c} \in [0, 1]$ controls the relative importance of
intermediate versus leaf-level supervision. This formulation allows the model
to propagate structural information throughout the taxonomy while maintaining
precision at the leaf level, preventing error accumulation across deeper
hierarchy layers.

\subsubsection{Auxiliary Semantic Consistency Loss}
To align model predictions with the semantic meaning of official tax code
definitions, we introduce an auxiliary task that enforces semantic consistency.
For each class label $y_{i}$, let $d(y_{i})$ denote its official description in
the tax code catalogue. A semantic encoder converts the pair ${y_{i},
d(y_{i})}$ into a latent representation, producing a reference semantic label
$z$. In parallel, the encoded product title is projected into a predicted
semantic label $\hat{z}$ by a lightweight linear layer. The loss is then
defined as:
\begin{equation}
  \mathcal{L}_{s} = \mathrm{cross\_entropy}(\hat{z}, z),
  \label{eqn:semantic-loss}
\end{equation}
This auxiliary supervision encourages predictions that are semantically
consistent with expert-verified tax code definitions, thereby reducing
misclassifications caused by surface-level textual similarity or ambiguous
descriptions. The two objectives are combined into the total training loss:
\begin{equation}
  \mathcal{L} = \omega_{s} \mathcal{L}_{c} + (1 - \omega_{s}) \mathcal{L}_{s},
  \label{eqn:total-loss}
\end{equation}
where the weight $\omega_{s} \in [0, 1]$ balance the contribution of structural
and semantic learning.

\subsubsection{Dataset Construction}
To support large-scale training, we construct four internal datasets that
together reflect the full tax code workflow illustrated in \Cref{fig:business}.
Each dataset corresponds to a distinct stage in the product-to-invoice life
cycle, covering both B2C and B2B operations. \Cref{tab:dataset-source}
summarizes their sources and cardinalities:

\begin{table}[!htb]
  \centering
  \caption{Internal datasets used for model training.}
  \label{tab:dataset-source}

  \begin{tabular}{llr}
    \toprule
    \textbf{Source}              & \textbf{Dataset}  & \textbf{Samples} \\
    \midrule
    Tmall Supermarket Stock      & Goods Registry    &     702{,}869 \\
    Business Event Records       & Knowledge Base    &     198{,}010 \\
    Invoice Validation Records   & Validation Record &     331{,}195 \\
    Domestic Output Invoice Pool & Invoice Archive   & 7{,}400{,}210 \\
    \bottomrule
  \end{tabular}
\end{table}

\begin{itemize}
  \item \textit{Goods Registry.} Collected from in-stock records of the Tmall
    Supermarket. When merchants register products on the e-commerce platform,
    the system logs aligned pairs of \textit{(product title, tax code)} or
    \textit{(product category, tax code)}. It captures the merchant-facing
    process of code assignment.
  \item \textit{Knowledge Base.} This manually curated repository is maintained
    by tax experts and periodically updated during business events. Whenever a
    product with a new title or category appears without an existing match,
    experts validate and insert the corresponding tax code entry.
  \item \textit{Validation Record.} Extracted from the invoice validation
    process, this dataset contains manual correction cases where automatically
    retrieved codes failed validation. These samples provide valuable
    supervision signals for identifying hard or ambiguous examples.
  \item \textit{Invoice Archive.} Collected from internal aggregated and
    de-identified invoice dataset that includes validated codes.
\end{itemize}

To ensure data integrity, all datasets undergo strict de-duplication based on
normalized product titles and tax code identifiers before partitioning into
training, validation, and test splits. Together, these corpora provide
comprehensive coverage of both B2C and B2B operations, forming a robust
foundation for hierarchical and semantic supervision.

\subsubsection{Training Pipeline}
\begin{figure}[t]
  \centering
  \includegraphics[width=\columnwidth]{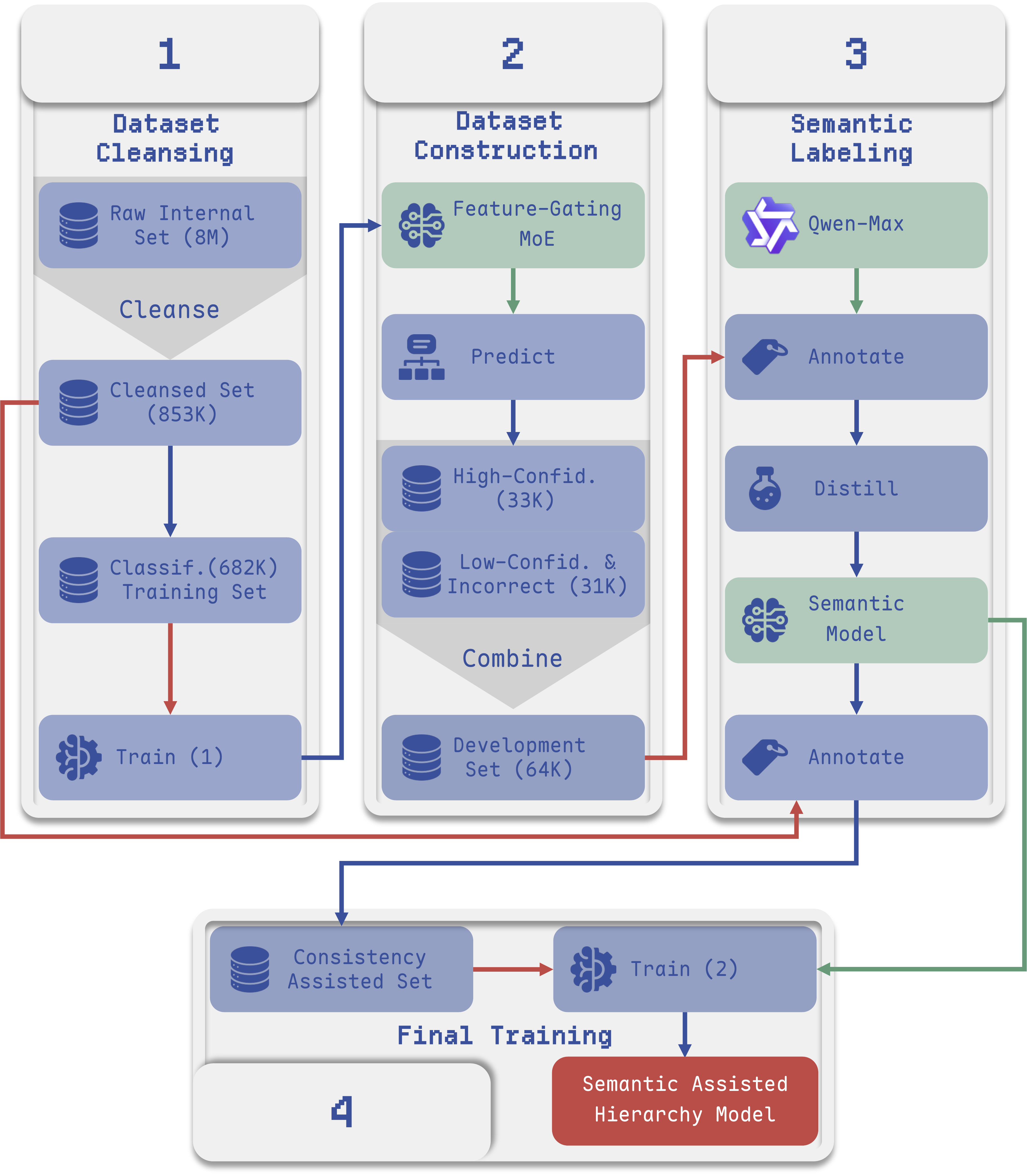}

  \caption{Four-stage data processing and training pipeline.}
  \label{fig:training-pipeline}
\end{figure}

To maximize data quality and training efficiency, we design a four-stage
pipeline that integrates dataset cleansing, semantic labeling, and joint
optimization. This process transforms the raw internal data into a high-quality
corpus suitable for hierarchical and semantic supervision. The overall workflow
is illustrated in \Cref{fig:training-pipeline}.

\paragraph{Stage 1 - Dataset Cleansing}
Starting from an 8M-sample raw internal set, we perform intra-category
clustering and rule-based filtering to remove mislabeled, duplicated, or
inconsistent entries, yielding a 682{,}207-sample cleansed set.

\begin{figure}[!htb]
  \centering
  \includegraphics[width=\columnwidth]{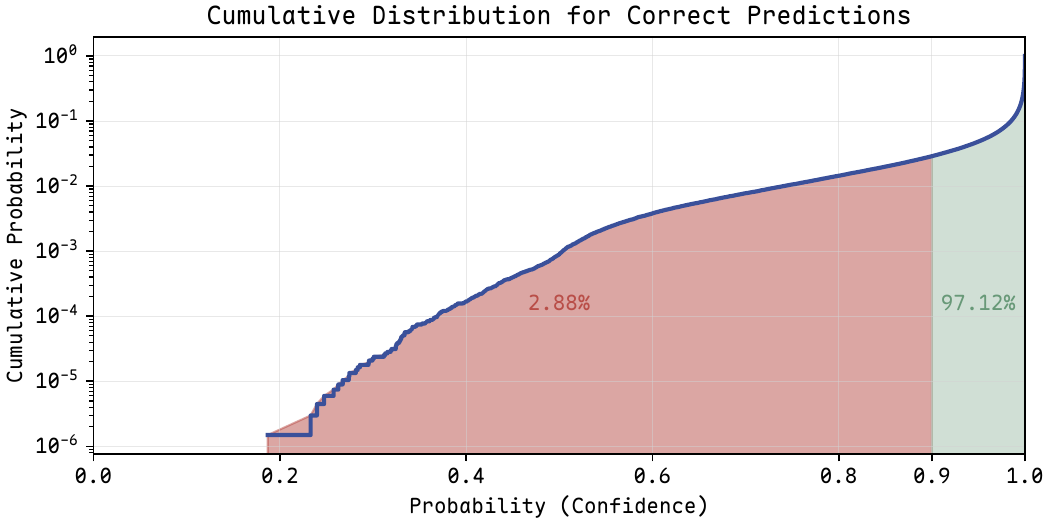}

  \caption{Log-scale Cumulative distribution of prediction confidence for
  correctly predicted samples. Only 2.88\% of predictions fall below 0.9
  confidence, motivating balanced sampling.}
  \label{fig:data-cdf}
\end{figure}

\paragraph{Stage 2 - Development Set Construction}
A preliminary hierarchical MoE model is trained on the cleansed set to assess
prediction confidence. Among 670{,}013 correctly predicted samples, fewer than
3\% fall below a confidence threshold of 0.9 (see \Cref{fig:data-cdf}). To
balance easy and hard cases, we randomly sample 5\% of high-confidence
positives and retain all low-confidence or incorrect samples, forming a
64{,}030-entry development set that supports semantic labeling and model
evaluation as shown in \Cref{tab:dataset-develop}.

\begin{table}[!htb]
  \centering
  \caption{Composition of the development dataset stratified by prediction
  correctness and confidence.}
  \label{tab:dataset-develop}

  \begin{tabular}{lrrrr}
    \toprule
    \textbf{Type}                 & \textbf{Samples}           & \textbf{Sampling}      & \textbf{Selected}         & \textbf{Consistency} \\
    \midrule
    \multirow{3}{*}{$P \geq 0.9$} & \multirow{3}{*}{650{,}713} & \multirow{3}{*}{5\%}   & \multirow{3}{*}{32{,}536} & \texttt{Y}: \hfill 72.3\% \\
                                  &                            &                        &                           & \texttt{N}: \hfill 21.5\% \\
                                  &                            &                        &                           & \texttt{-}: \hfill  6.2\% \\
    \midrule
    \multirow{2}{*}{$P < 0.9$}    & \multirow{2}{*}{19{,}300}  & \multirow{2}{*}{100\%} & \multirow{2}{*}{19{,}300} & \texttt{Y}: \hfill 59.6\% \\
    \multirow{2}{*}{Incorrect}    & \multirow{2}{*}{12{,}194}  & \multirow{2}{*}{100\%} & \multirow{2}{*}{12{,}194} & \texttt{N}: \hfill 38.2\% \\
                                  &                            &                        &                           & \texttt{-}: \hfill  2.2\% \\
    \midrule
    Total                         & 682{,}207                  & 64{,}030               & --                        & -- \\
    \bottomrule
  \end{tabular}
\end{table}

\begin{figure}[!htb]
  \hfill
  \begin{minipage}{0.9\columnwidth}
    \centering
    \includegraphics[width=\linewidth]{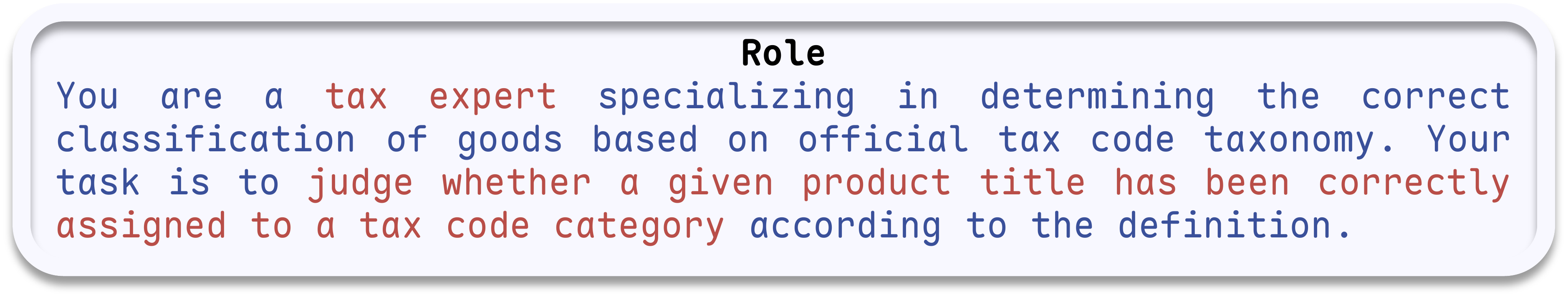}

    \caption{Role hinting the model to act as a tax expert and defines the task
    of semantic consistency judgement.}
    \label{fig:semantic-prompt-role}
  \end{minipage}
\end{figure}

\paragraph{Stage 3 - Semantic Labeling}
Entries in the development set are annotated with semantic consistency
judgments using Qwen-Max~\cite{qwen2024qwen}, a LLM acting as a domain expert.
We prompt it with structured JSON-based instructions to judge whether each
product title aligns with the definition of its assigned tax code. Each
annotation outputs one of three labels: \texttt{Y} (consistent), \texttt{N}
(inconsistent), or \texttt{-} (uncertain), together with a brief rationale.
These labels provide supervision for distilling a lightweight semantic judgment
model that mimics the LLM during large-scale training. Specifically:


\begin{enumerate}
  \item First, we provide the model with a role hint that constrains it to act
    as a professional tax expert. The role hint specifies the task of verifying
    whether the product title aligns with the official tax code definition, as
    shown in \Cref{fig:semantic-prompt-role}. This helps the model focus on
    semantic consistency within the tax domain.
\end{enumerate}

\begin{figure}[!htb]
  \hfill
  \begin{minipage}{0.9\columnwidth}
    \centering
    \includegraphics[width=\linewidth]{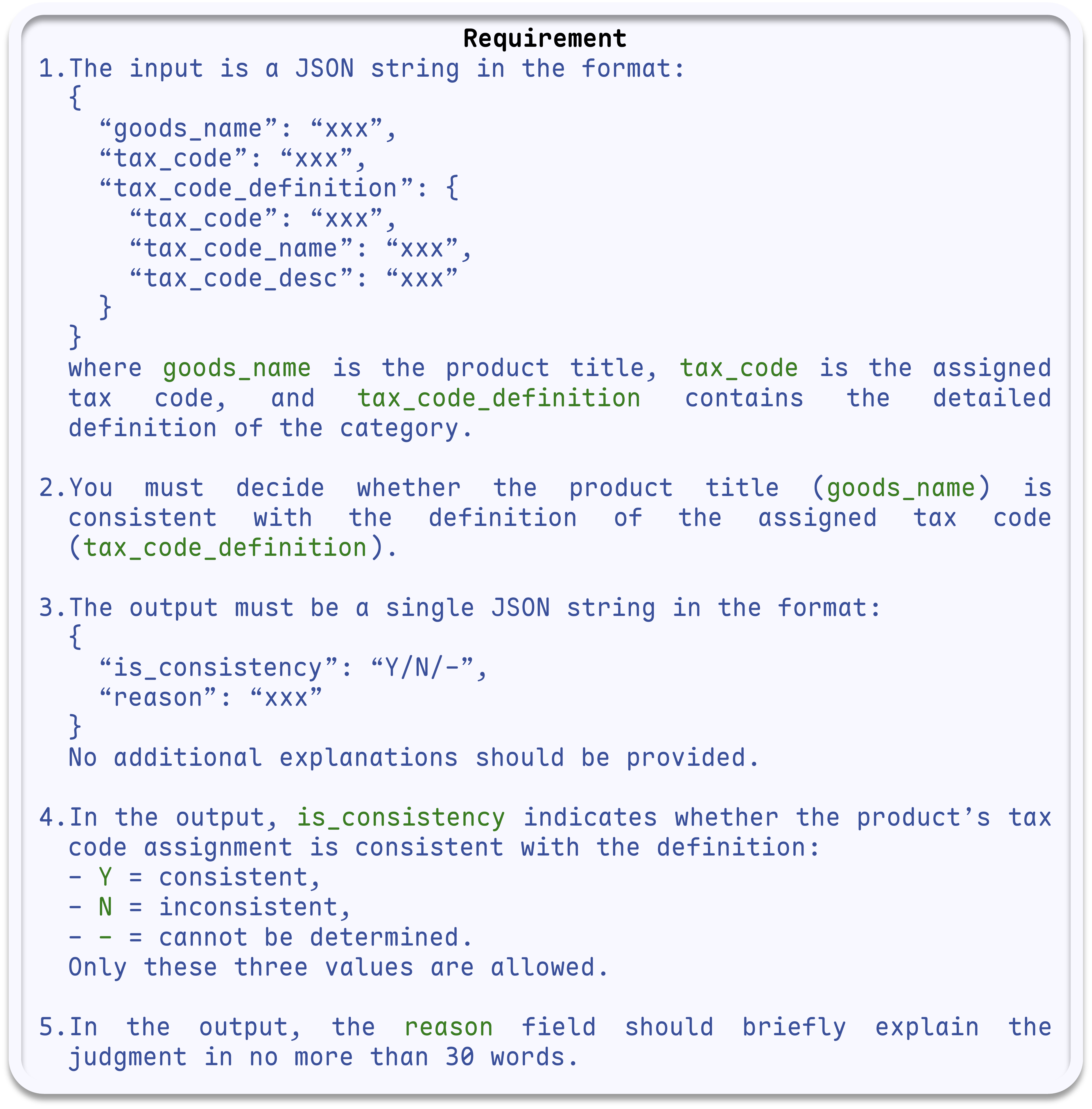}

    \caption{Structured requirements specifying input/output schemas, including
    field meanings and labeling rules, guiding the model's annotation process}
    \label{fig:semantic-prompt-requirement}
  \end{minipage}
\end{figure}

\begin{enumerate}
  \setcounter{enumi}{1}
  \item Next, we specify detailed structured requirements in JSON format that
    define both the input and output schemas. These schemas include field
    meanings and labeling rules, ensuring consistency in the labeling process,
    as shown in \Cref{fig:semantic-prompt-requirement}.
\end{enumerate}

\begin{figure}[!htb]
  \hfill
  \begin{minipage}{0.9\columnwidth}
    \centering
    \includegraphics[width=\linewidth]{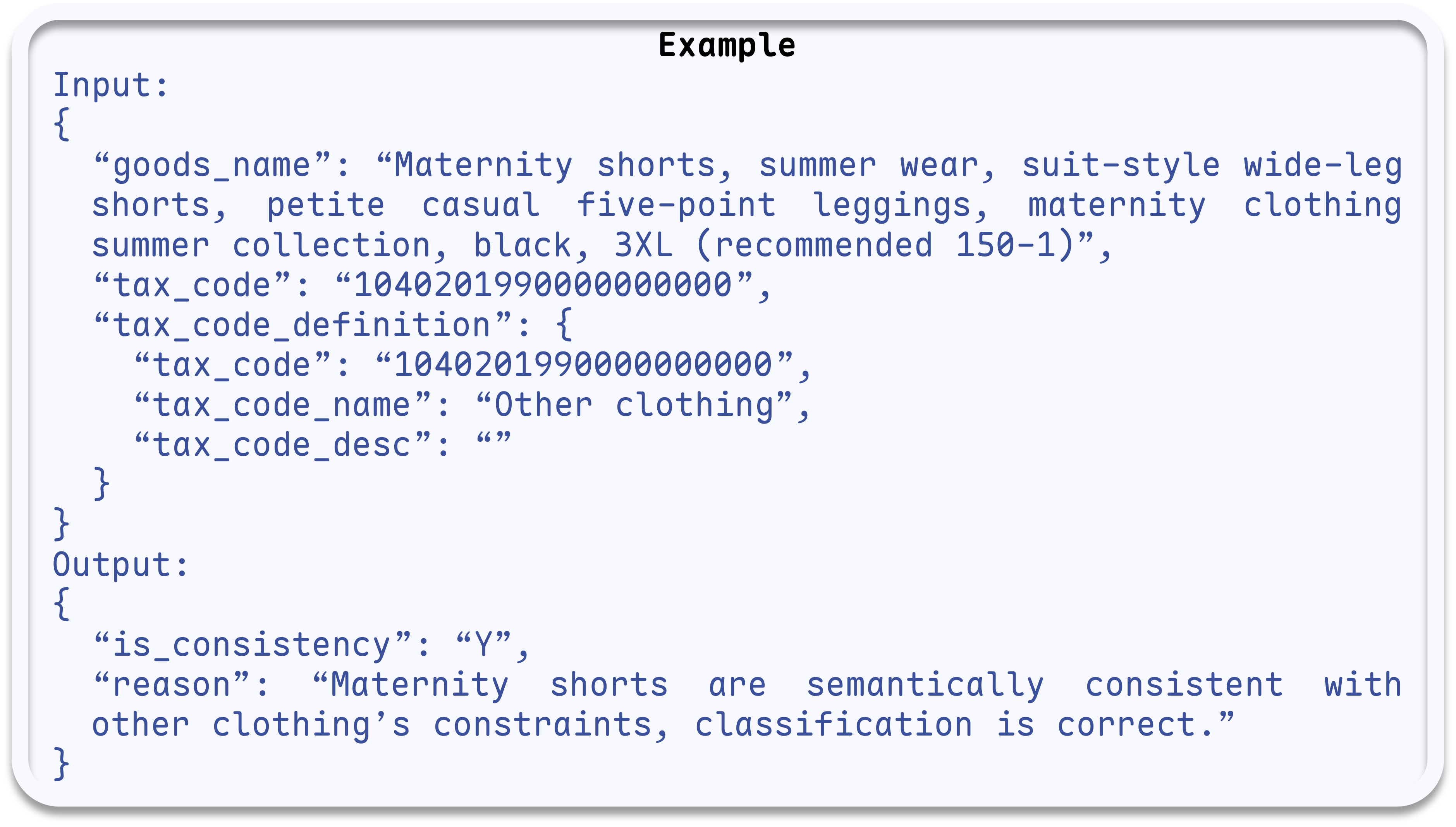}

    \caption{A few-shot example demonstrating the desired input-output behavior
    for semantic consistency judgment. This example helps guide the model's
    reasoning.}
    \label{fig:semantic-prompt-example}
  \end{minipage}
\end{figure}

\begin{enumerate}
  \setcounter{enumi}{2}
  \item Finally, we append three input-output examples as few-shot
    demonstrations to help guide the model's reasoning process, as shown in
    \Cref{fig:semantic-prompt-example}. For clarity, only one example is
    presented here, though the model is provided with multiple instances during
    training.
\end{enumerate}

Although the prompts are issued in Chinese to match our data, we present
English versions for readability. The resulting judgments are summarized in the
``Consistency'' column of \Cref{tab:dataset-develop} and distilled into the
semantic judgment model used in large-scale training.


\paragraph{Stage 4 - Final Consistency-Assisted Training}
The distilled semantic model is applied to the cleansed corpus to generate
consistency signals for all entries, producing a consistency-assisted training
set. The final hierarchical MoE model is then trained with the combined
objectives of hierarchical classification and semantic consistency (as defined
in \Cref{sec:model-training}). This joint optimization enforces both structural
fidelity to the taxonomic hierarchy and semantic alignment with official
definitions, resulting in a robust and interpretable tax code predictor.

\subsection{Business Integration}
\label{sec:business-integration}

\begin{figure}[!htb]
  \centering
  \includegraphics[width=\columnwidth]{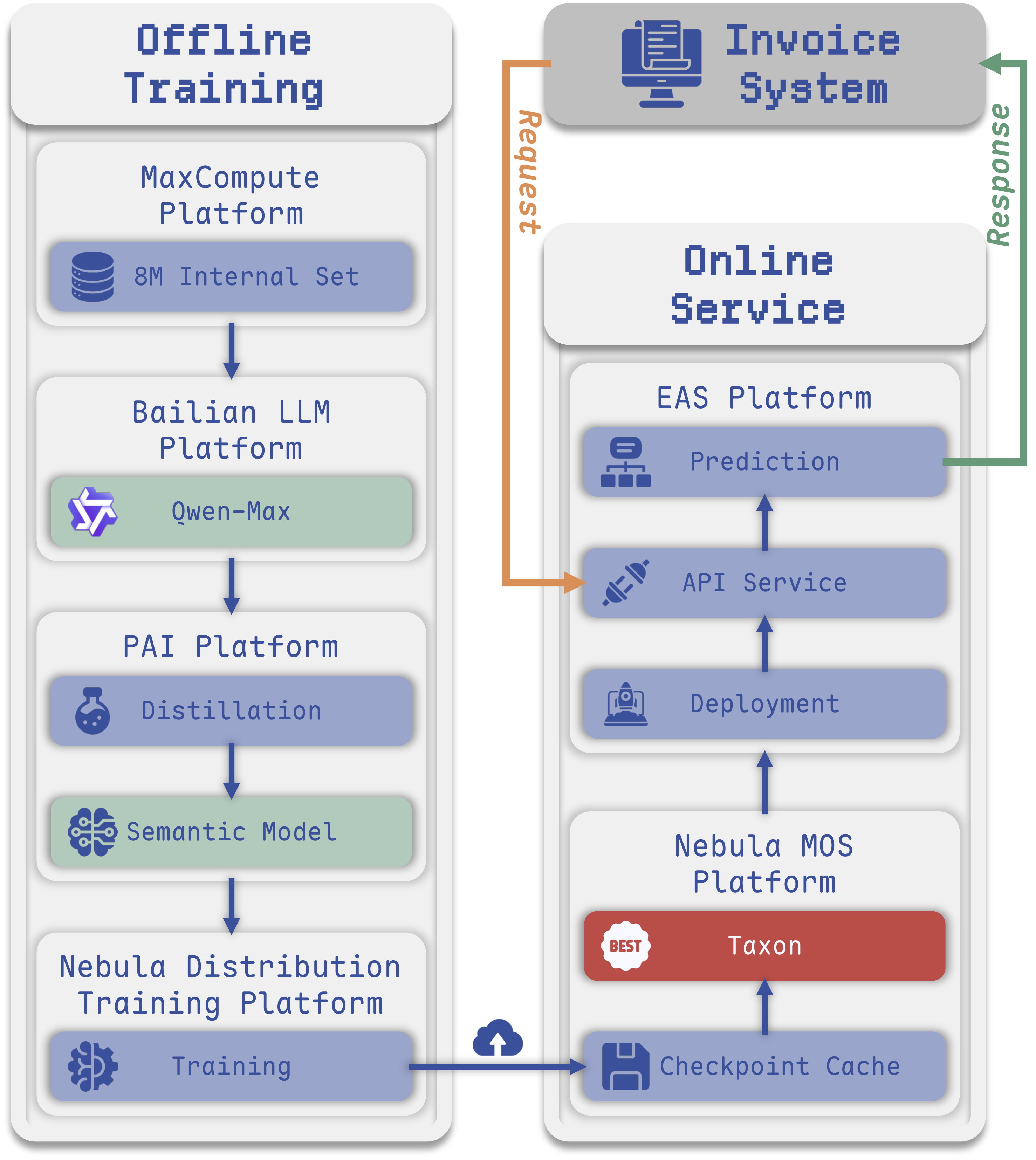}

  \caption{Integration of the proposed framework into Alibaba's enterprise tax
  service ecosystem.}
  \label{fig:integration}
\end{figure}

\Cref{fig:integration} illustrates the real-world deployment of \name{} within
Alibaba's ecosystem, forming a closed loop from data collection and semantic
distillation to real-time tax code prediction and validation in production.
Offline platforms handle data processing and model preparation, while online
platforms host the deployed service, enabling millions of daily tax code
queries from the invoice system.


\subsubsection{Offline Training}
We train \name{} on the full 8M-sample corpus using a single NVIDIA V100 GPU,
and end-to-end training takes about 3 days and 20 hours. The offline pipeline
consists of four stages across internal AI and data platforms:

\begin{enumerate}
  \item \textit{MaxCompute Platform} deduplicates and normalizes invoice
    issuance records and tax audit logs to construct the 8M internal dataset
    described in \Cref{sec:model-training}.
  \item \textit{Bailian LLM Platform} provides Qwen-Max inference APIs; a
    confidence-based strategy selects hard examples for consistency labeling.
  \item \textit{PAI Platform (Platform of Artificial Intelligence)} distills
    Qwen-Max annotations into a lightweight semantic model for corpus-wide
    semantic labeling.
  \item \textit{Nebula Distribution Training Platform} enables multi-node,
    multi-GPU distributed training for large-scale optimization.
\end{enumerate}

\subsubsection{Online Serving}
Trained checkpoints are uploaded to the \textit{Nebula Model Object Storage
(MOS) Platform} for unified version and metric management. The best-performing
model on the validation set is released as the production model,
\textit{\name{}}. It is deployed via the \textit{Elastic Algorithm Service
(EAS) Platform}, which encapsulates \name{} as a scalable API for real-time tax
code prediction in both product publishing and invoicing workflows. When a
request arrives from the \textit{Invoice System}--the group-wide invoicing
center--the EAS service performs prediction and validation, returning results
to the pipeline. The deployed system supports high concurrency and low latency,
sustaining about 30 queries per second with an average end-to-end response time
of about 90~ms handling about 5M requests daily and peaking up to 15M during
major business events.


This integration forms a continuous improvement loop: newly issued invoices and
expert-corrected samples are periodically fed back into MaxCompute for
incremental retraining, allowing \name{} to evolve with changing tax
regulations and product catalogs while maintaining accuracy, interpretability,
and compliance. Tax code catalogues evolve over time, but updates are typically
infrequent (often on a yearly cadence). When changes occur, we refresh the
internal knowledge base and retrain the model using newly issued invoices and
expert-corrected samples under the updated rules, ensuring \name{} stays
compliant as regulations and product catalogs evolve.

\section{Experiments}
\label{sec:experiments}
We conduct experiments to evaluate the effectiveness, generality, and
robustness of \name{}. Specifically, we examine: (i) whether expert-guided
semantic supervision improves accuracy across taxonomic depths; (ii) the
contribution of each architectural component; and (iii) how \name{} compares
with state-of-the-art hierarchical and label-semantic-aware baselines under
realistic business data conditions. We evaluate on both proprietary and public
datasets spanning diverse domains and hierarchy complexities. Unless otherwise
stated, all models use identical data splits and optimization settings. We
report results at both path- and leaf-level granularities, followed by
ablations and error analyses to quantify the impact of semantic alignment and
hierarchical modeling.


\subsection{Datasets}
\label{sec:main-dataset}
We conduct experiments on two datasets with distinct language and domain
characteristics to demonstrate the versatility of our approach:
\textit{TaxCode} (our proprietary dataset) \textit{and Web of Science
(WOS)}~\cite{kowsari2017hdltex}, as summarized in \Cref{tab:dataset}. We
exclude other widely used hierarchical text classification datasets such as
\textit{Reuters Corpus Volume I (RCV1)}~\cite{lewis2004rcv1} and \textit{The
New York Times Annotated Corpus (NYT)}~\cite{sandhaus2008nyt}, because they are
inherently multi-label, where each sample may correspond to multiple
intermediate or leaf nodes, whereas our formulation assumes a single
root-to-leaf path per instance.

\begin{table}[!htb]
  \centering
  \caption{Datasets for main and ablation experiments.}
  \label{tab:dataset}

  \resizebox{\columnwidth}{!}{%
  \begin{tabular}{llrrr}
    \toprule
    \textbf{Dataset}                & \textbf{Language}        & \textbf{Level}      & \textbf{Samples}           & \textbf{Split}           \\
    \midrule
    \multirow{3}{*}{TaxCode (Ours)} & \multirow{3}{*}{Chinese} & \multirow{3}{*}{10} & \multirow{3}{*}{852{,}758} & Train.: \hfill 545{,}765 \\
                                    &                          &                     &                            & Val.:   \hfill 136{,}441 \\
                                    &                          &                     &                            & Test.:  \hfill 170{,}551 \\
    \midrule
    \multirow{3}{*}{WOS}            & \multirow{3}{*}{English} & \multirow{3}{*}{2}  & \multirow{3}{*}{46{,}985}  & Train.: \hfill 30{,}070  \\
                                    &                          &                     &                            & Val.:    \hfill 7{,}518  \\
                                    &                          &                     &                            & Test.:   \hfill 9{,}397  \\
    \bottomrule
  \end{tabular}
  }
\end{table}

The \textit{TaxCode} dataset is a proprietary large-scale Chinese corpus
derived from e-commerce tax classification records, corresponding to the
cleansed set introduced in \Cref{fig:training-pipeline}. It reflects realistic
product and service labeling governed by national tax regulations, covering
diverse categories such as groceries, electronics, and logistics services.
After multi-stage filtering and de-duplication, the dataset comprises 852{,}758
instances labeled under a ten-level hierarchical taxonomy aligned with the
\textit{Goods and Services Tax Classification Catalogue}~\cite{sta2017goods}.
Each sample contains a product title, its hierarchical tax path, and structured
metadata (e.g., BU and OU codes), providing supervision for structural and
semantic modeling.

The \textit{Web of Science (WOS)} dataset~\cite{kowsari2017hdltex} is a widely
adopted English benchmark for hierarchical text classification. It consists of
46{,}985 research paper abstracts organized into a two-level taxonomy of
scientific disciplines. We include this dataset to assess cross-lingual and
cross-domain generalization. Unlike TaxCode, WOS contains well-formed textual
inputs and shallow hierarchies, making it a complementary test bed for
evaluating model robustness beyond noisy business data.

For both datasets, we partition them into training, validation, and test sets
in a 64/16/20 ratio to ensure balance across hierarchy levels, as summarized in
the ``Split'' column of \Cref{tab:dataset}. In the main experiments, we
randomly sample 5\% of the datasets for training, validation, and testing. For
the ablation experiments, we use the entire dataset to ensure comprehensive
evaluation.


\subsection{Metrics}
\label{sec:main-metrics}
We evaluate model performance using both \textit{macro} and \textit{micro}
$F_{1}$ scores at two granularities: \textit{path level} and \textit{leaf
level}. Metrics are derived by measuring the overlap between the predicted and
true label sets along the path:

\begin{equation}
  \begin{aligned}
    \mathrm{Precision} &= \frac{|\mathbf{\hat{y}} \cap \mathbf{y}|}{|\mathbf{\hat{y}}|},
    \quad
    \mathrm{Recall} = \frac{|\mathbf{\hat{y}} \cap \mathbf{y}|}{|\mathbf{y}|},
    \\
    F_{1} &= \frac{2 \times \mathrm{Precision} \times \mathrm{Recall}}{\mathrm{Precision} + \mathrm{Recall}}.
  \end{aligned}
  \label{eqn:metric}
\end{equation}

Macro $F_{1}$ measures the average performance across all categories.
Specifically, we first compute precision, recall, and $F_{1}$ for each label
and then average the scores across all 4{,}482 categories. In contrast, micro
$F_{1}$ aggregates all predictions and computes a single score over the entire
set of samples. The two perspectives jointly reflect both per-category
robustness and overall predictive accuracy.

\paragraph{Path-Level Evaluation}
At the path level, we treat each root-to-node path as a sequence of labels. We
compare the predicted path $\mathbf{\hat{y}}$ with the ground-truth path
$\mathbf{y}$ and compute $F_{1}$ from the overlap of their label sets.
Following the practice in prior hierarchical text classification
work~\cite{wang2022incorporating}, correctness is evaluated per node along the
path rather than requiring all ancestors to match, reflecting how well the
model captures taxonomy structure across depths.


\paragraph{Leaf-Level Evaluation}

At the leaf level, we focus on end-task correctness which evaluates whether the
final prediction matches the correct tax code, since taxation decisions operate
at the most specific level. For samples with partial annotations due to
incomplete metadata, we treat the deepest annotated node as the
\textit{effective leaf label}. If a prediction stops early (e.g., due to low
confidence), the farthest predicted node is used as the effective leaf
prediction. We compute $F_{1}$ over these effective leaf labels to quantify
end-task correctness.


\subsection{Baselines}
\label{sec:baselines}
We compare \name{} with representative hierarchical baselines including
HGCLR~\cite{wang2022incorporating}, HPT~\cite{wang2022hpt},
HILL~\cite{zhu2024hill}, HyILR~\cite{kumar2025hyilr}, and
LH-Mix~\cite{kong2025lhmix}. HGCLR and HILL are contrastive-learning approaches
that incorporate hierarchy-aware objectives; HPT is a prompt-tuning method that
aligns HTC with masked language modeling; HyILR models instance-specific local
hierarchical relations with hyperbolic geometry; and LH-Mix improves
hierarchical prompt tuning via local-hierarchy-guided Mixup. All methods are
trained and evaluated on the same data splits with identical evaluation
metrics, and we tune hyperparameters on the validation set.

We do not include a direct LLM in-context learning baseline, because the label
space contains over 4{,}000 leaf tax codes and the hierarchy is up to ten
levels deep, making na\"ive ``choose-from-all-labels'' prompting impractical;
retrieval- augmented prompting would introduce a retriever whose recall and
engineering choices materially affect end-to-end accuracy.

\subsection{Main Results}
\label{sec:main-results}
We evaluate our proposed framework on both public and internal business domain
datasets to assess its generality and practical effectiveness. The experiments
are designed to answer three key questions: (i) whether our expert-guided
knowledge distillation improves prediction accuracy across domains and
taxonomic depths; (ii) how semantic alignment and hierarchy modeling contribute
to overall performance; and (iii) how our approach compares with existing
hierarchy-aware and label-semantic-aware baselines under realistic data scales.
We report the main quantitative results on two representative datasets,
followed by detailed ablation and analysis studies.

\Cref{tab:main-results-taxcode} reports the main results on the
\textit{TaxCode} dataset at both path-level and leaf-level granularities. Among
the compared methods, our proposed \name{} framework achieves the overall best
performance, with the underlined values representing the highest results
without the final row. Compared the baselines, \name{} consistently yields
higher macro- and micro-level $F_{1}$ scores, demonstrating the effectiveness
of our semantically consistent and expert-guided framework in handling large
hierarchical label spaces.

\begin{table}[!htb]
  \centering
  \caption{$F_{1}$ across different methods on \textit{TaxCode} dataset.}
  \label{tab:main-results-taxcode}

  \begin{tabular}{lrrrr}
    \toprule
    \textbf{Method}                    & \multicolumn{2}{c}{\textbf{Path (\%)}} & \multicolumn{2}{c}{\textbf{Leaf (\%)}} \\
    \cmidrule{2-5}
                                       & \textbf{Macro}    & \textbf{Micro}     & \textbf{Macro}    & \textbf{Micro} \\
    \midrule
    HGCLR~\cite{wang2022incorporating} & 80.53             & \underline{90.18}  & 62.32             & 77.81             \\
    HPT~\cite{wang2022hpt}             & 73.13             & 85.09              & 40.71             & 62.79             \\
    HILL~\cite{zhu2024hill}            & 80.78             & 89.78              & 61.65             & 77.50             \\
    HyILR~\cite{kumar2025hyilr}        & 75.88             & 87.88              & 54.02             & 71.22             \\
    LH-Mix~\cite{kong2025lhmix}        & 79.84             & 89.24              & 63.73             & 76.49             \\
    \name{}                            & \underline{85.06} & 89.22              & \underline{78.30} & \underline{87.06} \\
    \name{} + RePath                   & \textbf{89.37}    & \textbf{93.16}     & \textbf{78.30}    & \textbf{87.06}    \\
    \bottomrule
  \end{tabular}
\end{table}

However, the path-level micro $F_{1}$ of \name{} (89.22) is slightly lower than
HILL (89.78) and LH-Mix (89.24). Examination of misclassified cases shows that
this gap mainly stems from path inconsistency rather than leaf errors--for
instance, the model may predict the correct leaf but include an extra
intermediate node. This indicates that while leaf predictions are accurate,
independent layer-wise classifiers can occasionally yield redundant or invalid
path nodes.


To address this issue, we conduct an additional experiment that reconstructs
the hierarchical path directly from the predicted leaf node, following the
corresponding ancestor chain in the taxonomy. We denote this variant as
``\textbf{\name{} + RePath}''. As shown in the last row of
\Cref{tab:main-results-taxcode}, this simple post-processing step substantially
improves the path-level macro and micro $F_{1}$ scores to 89.37 and 93.16,
respectively, while maintaining identical leaf-level results. The consistent
gain confirms that most residual path errors originate from internal-node
inconsistencies rather than semantic misunderstanding at the leaf level. By
leveraging the more reliable leaf predictions to reconstruct a coherent
root-to-leaf path, the RePath procedure effectively enhances the overall
structural correctness of hierarchical predictions.

To further analyze model behavior across different levels of label hierarchy,
we follow the statistics proposed in~\cite{kumar2025hyilr} and evaluate
performance with respect to \textit{path complexity}, which reflects the depth
of the hierarchical path defined by a specific tax code. This analysis provides
additional insight into how each model handles short versus long classification
paths. All test samples fall within path depths from 2 to 6, with 11, 17, 176,
2730, and 477 instances respectively for each depth.

\begin{figure}[!htb]
  \centering
  \subfloat[Path-Level Macro $F_{1}$]{
    \includegraphics[width=0.96\columnwidth]{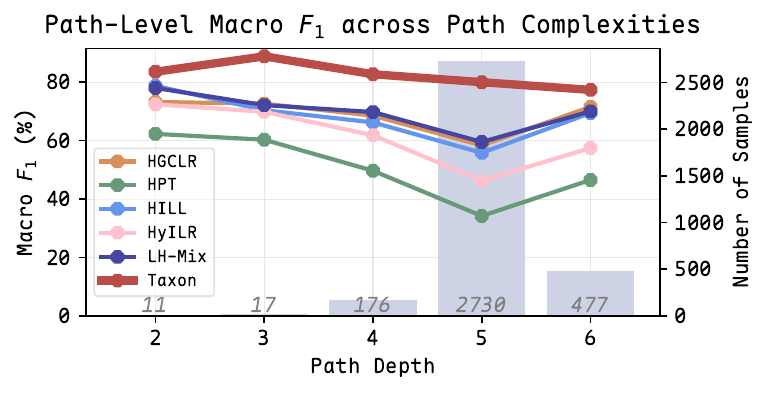}
    \label{fig:path-macro-f1}
  }
  \vfill
  \subfloat[Path-Level Micro $F_{1}$]{
    \includegraphics[width=0.96\columnwidth]{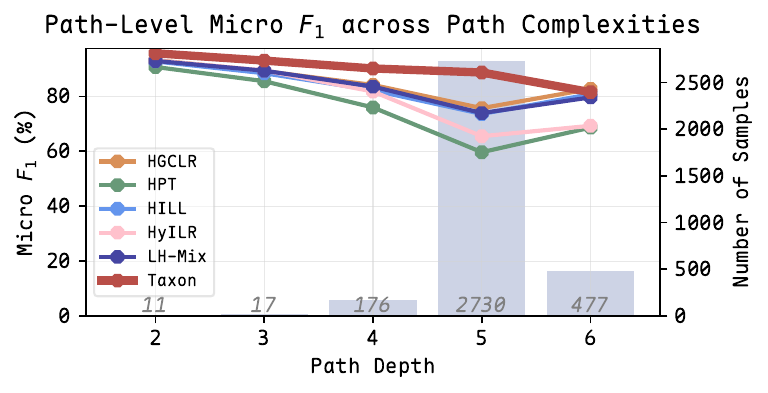}
    \label{fig:path-micro-f1}
  }

  \caption{Model performance at different levels of path complexity. Bars
  indicate the number of test samples at each depth.}
  \label{fig:path-f1}
\end{figure}

At the path level (\Cref{fig:path-f1}), both macro and micro $F_{1}$ scores of
\name{} remain consistently higher than those of all baselines across all
depths, demonstrating superior stability and robustness when predicting at
varying path complexities. Notably, all methods exhibit similar performance
trends as path complexity increases, suggesting that deeper and structurally
richer taxonomies remain more challenging for hierarchical prediction models.

\begin{figure}[!htb]
  \centering
  \subfloat[Leaf-Level Macro $F_{1}$]{
    \includegraphics[width=0.96\columnwidth]{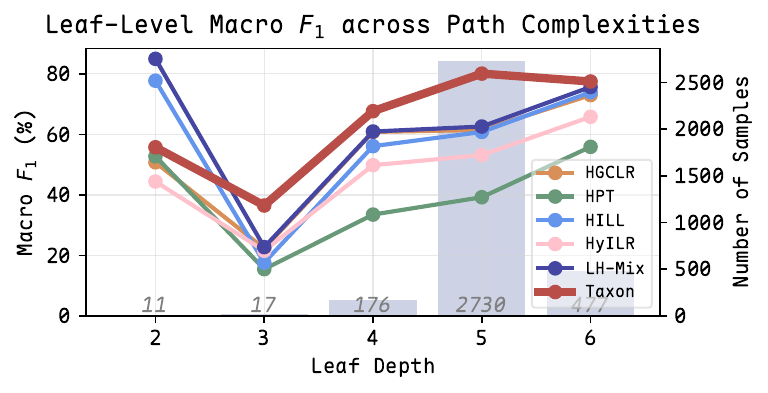}
    \label{fig:leaf-macro-f1}
  }
  \vfill
  \subfloat[Leaf-Level Micro $F_{1}$]{
    \includegraphics[width=0.96\columnwidth]{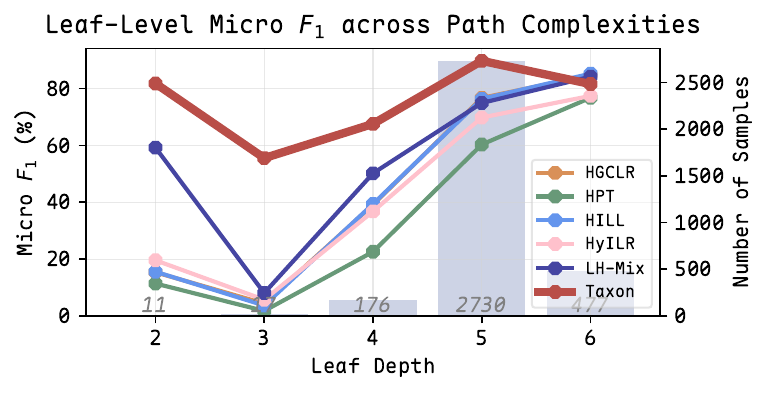}
    \label{fig:leaf-micro-f1}
  }

  \caption{Model performance at different levels of path complexity. Bars
  indicate the number of test samples at each depth.}
  \label{fig:leaf-f1}
\end{figure}

At the leaf level (\Cref{fig:leaf-f1}), \name{} shows slightly lower macro
$F_{1}$ than LH-Mix and HILL at depth 2, and lower micro $F_{1}$ than LH-Mix
and HILL at depth 6, but surpasses all other baselines at the remaining depths.
Notably, at depth 5 where the majority of samples reside, \name{} achieves the
highest macro and micro $F_{1}$ among all methods.

Beyond the proprietary \textit{TaxCode} dataset, we further evaluate our
framework on the public \textit{WOS} benchmark to examine \name{}'s
generalization capability under a shallower two-level hierarchy. The results
are reported in \Cref{tab:main-results-wos}, where we present both
\textbf{\name{}} and \textbf{\name{} + RePath} for consistency with the
previous experiment. Since the WOS taxonomy contains only two levels, the
RePath correction provides negligible improvement, as reflected by the
identical scores at the first two decimal places. Nonetheless, \name{} achieves
strong performance across all metrics, recording 86.74 macro $F_{1}$ and 93.67
micro $F_{1}$ at the path level, and 86.06 macro $F_{1}$ and 87.68 micro
$F_{1}$ at the leaf level. Our proposed method surpasses or matches
state-of-the-art baselines such as HGCLR, HILL, and LH-Mix. These results
confirm that \name{} generalizes effectively beyond the tax domain and remains
competitive even without domain-specific semantic supervision.


\begin{table}[!b]
  \centering
  \caption{$F_{1}$ across different methods on \textit{WOS} dataset.}
  \label{tab:main-results-wos}

  \begin{tabular}{lrrrr}
    \toprule
    \textbf{Method}                    & \multicolumn{2}{c}{\textbf{Path (\%)}} & \multicolumn{2}{c}{\textbf{Leaf (\%)}} \\
    \cmidrule{2-5}
                                       & \textbf{Macro}    & \textbf{Micro}     & \textbf{Macro}    & \textbf{Micro} \\
    \midrule
    HGCLR~\cite{wang2022incorporating} & \textbf{86.89}    & 91.23              & 85.84             & \textbf{88.08} \\
    HPT~\cite{wang2022hpt}             & 86.40             & 90.70              & 85.34             & 87.63          \\
    HILL~\cite{zhu2024hill}            & 86.19             & 90.68              & 85.40             & 87.77          \\
    HyILR~\cite{kumar2025hyilr}        & 86.22             & 90.74              & 85.07             & 87.56          \\
    LH-Mix~\cite{kong2025lhmix}        & 86.08             & 90.35              & 85.14             & 87.24          \\
    \name{}                            & 86.74             & \textbf{93.67}     & \textbf{86.06}    & 87.68          \\
    \name{} + RePath                   & 86.74             & \textbf{93.67}     & \textbf{86.06}    & 87.68          \\
    \bottomrule
  \end{tabular}
\end{table}

\subsection{Ablation Study}
\label{sec:ablation-study}
To systematically assess the contribution of each component in our proposed
framework, we design a progressive series of ablation experiments. Starting
from a minimal setup that combines a text encoder with a linear classification
head, we incrementally integrate additional modules to evaluate their
individual effects:

\begin{enumerate}
  \item Text encoder backbone;
  \item Category name feature fusion;
  \item Feature-gating MoE substitution;
  \item Hierarchical supervision;
  \item Semantic consistency auxiliary task.
\end{enumerate}

This stepwise design allows us to isolate where each enhancement contributes
most to performance improvements.

\subsubsection{Text Encoder Choice}
\label{sec:ablation-encoder}
We begin by comparing different text encoders to identify the most effective
backbone for representing product descriptions. Experiments are conducted on
the \textit{Goods Registry} dataset from the Tmall supermarket stock. As shown in
\Cref{tab:ablation-encoder}, transformer-based encoders significantly
outperform the convolutional baseline. Among them, BERT achieves the highest
accuracy (93.28\%), slightly ahead of XLNet (93.21\%), and clearly surpassing
TextCNN (91.26\%). This demonstrates that contextualized representations from
pre-trained language models
\begin{wraptable}[8]{r}{0.5\linewidth}
  \centering
  \footnotesize  
  \caption{Comparison of text encoders on \textit{Goods Registry}.}
  \label{tab:ablation-encoder}

  \begin{tabular}{lr}
    \toprule
    \textbf{Model} & \textbf{Accuracy} \\
    \midrule
    TextCNN        & 91.26\%           \\
    XLNet          & 93.21\%           \\
    BERT           & \textbf{93.28\%}  \\
    \bottomrule
  \end{tabular}
\end{wraptable}
capture the nuanced semantics in short and noisy product titles more
effectively than shallow convolutional encoders. Based on this result, BERT is
selected as the major encoder in all subsequent ablation studies.

\subsubsection{Category Name Fusion}
\label{sec:ablation-category-fusion}
We then inspect whether incorporating structured business features can enhance
classification accuracy. Since most business attributes (e.g., business unit or
organization unit codes) lack intrinsic semantics, we focus on the
\textit{category name} as the only semantically meaningful feature. This
feature is integrated into the model in one-hot or textual forms. To ensure
that the effectiveness of the category feature is not confined to a single
backbone, the one-hot variant is applied to all three encoders (TextCNN, XLNet,
and BERT), while the textual variant is evaluated only on BERT--the
best-performing text encoder from the previous study.

As shown in \Cref{tab:ablation-category-fusion}, incorporating the category
name in one-hot form consistently improves performance across all encoders and
datasets, demonstrating that structured categorical context complements other
input signals such as BU/OU codes and product titles. By comparing the
configurations with one-hot category names, we can again confirm that BERT
remains the most effective backbone under our experimental setup. Furthermore,
substituting the one-hot encoding with a textual form yields additional gains
(e.g., from 90.05\% to 90.64\% on the multi-source setting). These results
suggest that embedding category names facilitates stronger alignment between
input features and tax code labels.

\begin{table}[!htb]
  \centering
  \caption{Comparison of category name fusion strategies.}
  \label{tab:ablation-category-fusion}

  \resizebox{\columnwidth}{!}{%
  \begin{tabular}{lrrrr}
    \toprule
    \textbf{Configuration} & \textbf{Goods}    & \textbf{Validation} & \textbf{Invoice} & \textbf{Multi-Source} \\
                           & \textbf{Registry} & \textbf{Record}     & \textbf{Archive} & \textbf{(All Three)}  \\
    \midrule
    TextCNN + One-Hot      & 96.16\%           & 88.51\%             & 85.80\%          & 86.77\%               \\
    XLNet + One-Hot        & 94.32\%           & 89.71\%             & 89.45\%          & 89.89\%               \\
    BERT + One-Hot         & 94.48\%           & 89.91\%             & 89.60\%          & 90.05\%               \\
    BERT + Textual         & \textbf{96.61\%}  & \textbf{91.54\%}    & \textbf{90.04\%} & \textbf{90.64\%}      \\
    \bottomrule
  \end{tabular}
  }
\end{table}

\subsubsection{Classifier Choice}
\label{sec:ablation-classifier}
Building on the previous setup with BERT and fused category features, we
further replace the linear classification head with our proposed feature-gating
MoE layer. This module dynamically routes feature representations to
specialized expert submodules based on feature semantics, enabling more
adaptive decision boundaries that better capture the complex relationships
between different feature types.

As shown in \Cref{tab:ablation-classifier}, the feature-gating MoE classifier
consistently outperforms the linear classifier across all datasets, achieving
the largest gain on the \textit{Validation Record} set, with a +1.25\% absolute
improvement. These results highlight the effectiveness of adaptive expert
routing, which models inter-feature specialization, improving both
classification accuracy and generalization across diverse and heterogeneous
business domains.

\begin{table}[!htb]
  \centering
  \caption{Comparison of classification heads.}
  \label{tab:ablation-classifier}

  \resizebox{\columnwidth}{!}{%
  \begin{tabular}{lrrrr}
    \toprule
    \textbf{Configuration} & \textbf{Goods}    & \textbf{Validation} & \textbf{Invoice} & \textbf{Multi-Source} \\
                           & \textbf{Registry} & \textbf{Record}     & \textbf{Archive} & \textbf{(All Three)}  \\
    \midrule
    BERT + Linear          & 96.61\%           & 91.54\%             & 90.04\%          & 90.64\%               \\
    BERT + MoE             & \textbf{96.61\%}  & \textbf{92.79\%}    & \textbf{90.11\%} & \textbf{90.77\%}      \\
    \bottomrule
  \end{tabular}
  }
\end{table}

\subsubsection{Hierarchical Supervision}
\label{sec:ablation-hierarchical}
To investigate the impact of hierarchical classification, we introduce a
multi-level loss that supervises intermediate-level predictions along the
taxonomic path, with a loss weight of $\omega_{c} = 0.2$ (see
\Cref{eqn:classification-loss} and \Cref{fig:model-training}). As a comparison,
we also explore a sequence-generation approach that produces intermediate-level
labels.

As shown in \Cref{tab:ablation-hierarchical}, incorporating the multi-level
loss consistently improves accuracy across all datasets. This demonstrates that
hierarchical supervision enables the model to better capture the underlying
structure of the tax code hierarchy. In contrast, replacing ground-truth labels
with sequentially generated labels results in a slight reduction in accuracy,
highlighting the importance of high-quality hierarchical signals for effective
supervision.

\begin{table}[!b]
  \centering
  \caption{Effectiveness of hierarchical classification.}
  \label{tab:ablation-hierarchical}

  \resizebox{\columnwidth}{!}{%
  \begin{tabular}{lrrrr}
    \toprule
    \textbf{Configuration} & \textbf{Goods}                    & \textbf{Validation}               & \textbf{Invoice}                  & \textbf{Multi-Source} \\
                           & \textbf{Registry}                 & \textbf{Record}                   & \textbf{Archive}                  & \textbf{(All Three)}  \\
    \midrule
    BERT + MoE             & 96.61\%                           & 92.79\%                           & 90.11\%                           & 90.77\%                           \\
    \midrule
    BERT + MoE             & \multirow{2}{*}{96.40\%}          & \multirow{2}{*}{91.88\%}          & \multirow{2}{*}{89.43\%}          & \multirow{2}{*}{90.11\%}          \\
    + Sequential           &                                   &                                   &                                   &                                   \\
    \midrule
    BERT + MoE             & \multirow{2}{*}{\textbf{96.69\%}} & \multirow{2}{*}{\textbf{93.04\%}} & \multirow{2}{*}{\textbf{90.31\%}} & \multirow{2}{*}{\textbf{90.97\%}} \\
    + Hierarchical         &                                   &                                   &                                   &                                   \\
    \bottomrule
  \end{tabular}
  }
\end{table}

\subsubsection{Semantic Consistency}
\label{sec:ablation-semantic}
Finally, we evaluate the effect of the proposed semantic consistency auxiliary
task, which explicitly encourages predictions to align with the official
definitions of tax codes. Due to the additional computational cost, this
ablation is conducted on the smaller development set (as defined in
\Cref{tab:dataset-develop} and illustrated by block~2 in
\Cref{fig:training-pipeline}).


As shown in \Cref{tab:ablation-semantic}, introducing the semantic loss with a
weight of $\omega_{s}=0.2$ (see \Cref{eqn:semantic-loss} and
\Cref{fig:model-training}) yields consistent yet moderate improvements across
datasets, except for the \textit{Validation Record} set, where results remain
comparable. We further test a variant that replaces the distilled semantic
annotation model (block~3 of \Cref{fig:training-pipeline}) with an untrained
Qwen-2.5-32B~\cite{qwen2024qwen} model for direct semantic labeling, which
achieves smaller gains, confirming the importance of domain adaptation in the
distilled model.


Although the quantitative improvement appears limited, the qualitative effects
are more substantial. We summarize three main reasons: (i) our semantic model
corrects noisy labels in the ground truth, so while apparent agreement with the
original annotations decreases, true performance improves; (ii) it effectively
fixes most knowledge-level misclassifications, where the predicted tax code
better matches the product's meaning even if it differs from the noisy label;
and (iii) residual errors arise partly from imperfect semantic labels--the
LLM's own labeling inaccuracies occasionally misguide training, leaving a small
portion of hard cases unresolved. Together, these findings indicate that
explicit semantic alignment enhances interpretability and robustness, refining
the model's understanding of the relationship between product text and formal
tax code definitions.


\begin{table}[!htb]
  \centering
  \caption{Effectiveness of semantic consistency.}
  \label{tab:ablation-semantic}

  \resizebox{\columnwidth}{!}{%
  \begin{tabular}{lrrrr}
    \toprule
    \textbf{Configuration} & \textbf{Goods}                    & \textbf{Validation}               & \textbf{Invoice}                  & \textbf{Multi-Source} \\
                           & \textbf{Registry}                 & \textbf{Record}                   & \textbf{Archive}                  & \textbf{(All Three)}  \\
    \midrule
    BERT + MoE             & \multirow{2}{*}{98.30\%}          & \multirow{2}{*}{\textbf{97.50\%}} & \multirow{2}{*}{92.13\%}          & \multirow{2}{*}{94.71\%}          \\
    + Hierarchical         &                                   &                                   &                                   &                                   \\
    \midrule
    BERT + MoE             & \multirow{3}{*}{98.51\%}          & \multirow{3}{*}{96.83\%}          & \multirow{3}{*}{92.35\%}          & \multirow{3}{*}{94.89\%}          \\
    + Hierarchical         &                                   &                                   &                                   &                                   \\
    + Qwen2.5-32B          &                                   &                                   &                                   &                                   \\
    \midrule
    BERT + MoE             & \multirow{3}{*}{\textbf{98.69\%}} & \multirow{3}{*}{96.82\%}          & \multirow{3}{*}{\textbf{92.47\%}} & \multirow{3}{*}{\textbf{94.92\%}} \\
    + Hierarchical         &                                   &                                   &                                   &                                   \\
    + Semantic             &                                   &                                   &                                   &                                   \\
    \bottomrule
  \end{tabular}
  }
\end{table}

Across all ablation stages, we observe a cumulative improvement trend: (i)
pre-trained Transformer backbones, particularly BERT, offer stronger contextual
representations than shallow CNNs; (ii) incorporating category name features
adds structured semantic cues that complement other business attributes; (iii)
the feature-gating MoE improves cross-domain generalization through adaptive
expert specialization; (iv) hierarchical supervision enables structurally
consistent multi-level predictions; and (v) the semantic consistency task
further refines alignment with official definitions. Together, these modules
contribute distinct yet complementary gains, culminating in a semantically and
hierarchically consistent tax code predictor.


\section{Related Work}
\label{sec:related}
Tax code prediction can be viewed as a special case of hierarchical text
classification (HTC), where the label space follows a multi-level taxonomic
structure. This section reviews related work from two perspectives: (i) general
methods for HTC, and (ii) studies specifically targeting tax or similar
prediction or classification systems.

\subsection{Hierarchical Text Classification}
\label{sec:htc}
HTC aims to assign each text to one or more labels organized in a hierarchical
taxonomy. Prior work can be broadly categorized along two dimensions: (i) the
\textit{task form} (sequence generation vs.\ classification), and (ii) the
\textit{methodological emphasis} (hierarchy-aware, label-semantic-aware,
hybrid, and prompt/LLM-based). Generative methods based on CNNs or RNNs
suffered from exposure bias and error
propagation~\cite{peng2018large,shimura2018hft,banerjee2019hierarchical,wehrmann2018hierarchical},
and generally underperformed classification approaches in fully supervised
settings~\cite{zhou2020hierarchy,chen2021hierarchy,yu2022constrained}. With the
rise of prompt tuning and LLMs, generative formulations have shown strong
few-shot performance. However, when training data are sufficient, the most
competitive results are still achieved by hierarchy-aware and
label-semantic-aware methods.


\paragraph{Sequence Generation Methods}
Sequence generation approaches recast hierarchical classification as a path
prediction task. SGM~\cite{yang2018sgm} introduced a decoder-based architecture
to capture label dependencies for multi-label classification.
Seq2Tree~\cite{yu2022constrained} modeled HTC as a sequence-to-tree problem
with constrained decoding for label consistency, while
PAAM-HiA-T5~\cite{huang2022exploring} proposed a hierarchy-aware
T5~\cite{raffel2020exploring} with path-adaptive attention for level/path
dependency modeling. UMP-MG~\cite{ning2023ump} leveraged unidirectional message
passing to respect tree directionality, and HiDEC~\cite{im2023hierarchical}
achieved compact hierarchy-aware decoding with fewer parameters. Recent work
such as HiGen~\cite{jain2024higen} added dynamic, level-guided losses and
task-specific pre-training to mitigate data imbalance. Despite progress, error
propagation along decoding paths remains a key challenge, especially under deep
hierarchies.


\paragraph{Hierarchy-Aware and Label-Semantic-Aware Methods}
A large body of research explicitly models label hierarchies and semantics.
Hierarchy-aware models such as AttentionXML~\cite{you2019attentionxml},
HTrans~\cite{banerjee2019hierarchical}, and HiAGM~\cite{zhou2020hierarchy}
capture structural dependencies via hierarchical attention, recurrent transfer
learning, or GCN/TreeLSTM encoders. In parallel, label-semantic-aware methods
leverage label names and definitions: HTCInfoMax~\cite{deng2021htcinfomax}
maximizes text--label mutual information, while
HiMatch~\cite{chen2021hierarchy} formulates HTC as text--label semantic
matching in a shared embedding space. These lines are increasingly unified in
hybrid models such as HBGL~\cite{jiang2022exploiting},
HGCLR~\cite{wang2022incorporating}, and K-HTC~\cite{liu2023enhancing},
integrating hierarchy-guided contrastive learning with external knowledge
graphs. Recent variants (HiTIN~\cite{zhu2023hitin},
HJCL~\cite{yu2023instances}, HALB~\cite{zhang2024hierarchy},
HBM~\cite{kim2024hierarchy}, HILL~\cite{zhu2024hill},
HiAdv~\cite{wang2024utilizing}) further extend this paradigm with
structure-entropy encoders, supervised contrastive learning,
asymmetric/adaptive losses, and local hierarchy modeling. Other work explores
alternative geometries and negative sampling, including hyperbolic embeddings
(HyILR~\cite{kumar2025hyilr}) and hierarchical ranking
(HiSR~\cite{zhou2025novel}), yielding incremental but consistent gains.


\paragraph{Prompt-Tuning and LLM-Based Methods}
The emergence of large pre-trained models and parameter-efficient tuning has
brought new perspectives to HTC. HPT~\cite{wang2022hpt} proposed
hierarchy-aware prompt tuning to align multi-label classification with masked
language modeling objectives, while HierVerb~\cite{ji2023hierarchical} designed
path-constrained verbalizers and hierarchical contrastive learning for few-shot
HTC. NERHTC~\cite{cai2024ner} reformulated HTC as a named-entity-recognition
task with CRF-based path consistency. Another recent advance,
LH-Mix~\cite{kong2025lhmix}, enhances prompt tuning for HTC by integrating
text-specific local hierarchies into manually designed depth-level prompts. To
further capture implicit correlations among sibling categories, it applies a
Mixup strategy guided by local hierarchy correlation, substantially improving
robustness across multiple benchmark datasets. LLM-based extensions further
broadened this frontier. Retrieval-style ICL~\cite{chen2024retrieval} applied
in-context learning with hierarchical retrieval databases;
DPT~\cite{xiong2024dual} introduced dual prompt tuning with cross-layer
contrastive objectives; and TELEClass~\cite{zhang2025teleclass},
TTC~\cite{chen2025leveraging}, and KG-HTC~\cite{zang2025kg} combined LLMs with
taxonomy enrichment, multimodal consistency, and knowledge graph–based RAG
mechanisms for zero- or few-shot HTC. These models excel under low-resource
conditions but remain difficult to scale efficiently to large taxonomies.

To summarize, HTC research spans structural, semantic, hybrid, and LLM-based
paradigms, with the current state-of-the-art in hybrid hierarchy- and
label-semantic-aware frameworks. However, existing models often encode
hierarchy with explicit structure encoders (e.g., GCN, TreeLSTM) and model
label semantics via contrastive alignment. We instead propose a
\textit{hierarchical feature-gating MoE} framework, where hierarchical
relations are implicitly learned through expert routing and semantic relations
are strengthened by a judgment task. This architecture offers a new way to
balance global structural consistency with fine-grained semantic
discrimination.


\subsection{Tax Code and Harmonized System Classification}
\label{sec:tax-code-and-hs}
Tax code and Harmonized System (HS) classification constitute real-world
instances of HTC problems, where labels form deep taxonomies used for customs,
taxation, and trade management. The HS code system, established by the World
Customs Organization (WCO), defines a six-digit international standard extended
by local administrations (e.g., 10-digit CN codes in China and the U.S.). Such
multi-level taxonomies pose similar challenges of fine-grained disambiguation,
label imbalance, and semantic overlap observed in HTC.

Luppes et al.~\cite{luppes2019classifying} pioneered CNN-based classification
for HS codes, demonstrating strong performance ($F_{1}$ score up to 0.92 at the
HS-2 level) using domain-specific embeddings. Zhao et al.~\cite{zhao2020tax}
addressed value-added tax (VAT) classification in Chinese invoices via a
Heterogeneous Directed Graph Attention Network (HDGAT), improving accuracy by
leveraging relational dependencies among invoice items. Subsequent works
expanded model complexity and modality. Liao et al.~\cite{liao2024enhanced}
integrated ERNIE, BiLSTM, and multi-scale attention mechanisms to handle
semi-structured Chinese product descriptions with domain-specific terminology.
Amel et al.~\cite{amel2023multimodal} introduced a multimodal deep learning
framework combining text and image representations from e-commerce
declarations, achieving top-5 accuracy exceeding 98\%. Shubham et
al.~\cite{shubham2023ensemble} developed a hierarchical conditional BERT
classifier and a knowledge-graph–based auditing mechanism for HS code
assignment, while Anggoro et al.~\cite{anggoro2025harmonized} applied
contrastive learning with Sentence-BERT and Multiple Negative Ranking loss to
improve embedding quality and downstream classification across international
customs datasets.

Collectively, these studies highlight the persistent challenges of
high-dimensional, hierarchical label spaces and semantic ambiguity in product
descriptions. However, existing solutions are largely classification-based and
rely on handcrafted hierarchy modeling. Our work extends this line by
introducing a unified MoE-based framework that jointly learns hierarchical
structure and semantic alignment, offering a scalable, end-to-end approach to
national tax code prediction.

\section{Conclusion}
\label{sec:conclusion}
In this work, we addressed the challenge of hierarchical tax code prediction, a
mission-critical yet previously underexplored task in large-scale invoicing
systems. We introduced \name{}, a semantically aligned and expert-guided
prediction framework that unifies hierarchical modeling and semantic reasoning.
By combining a feature-gating MoE classifier for structure-aware routing with
an LLM-distilled semantic consistency task, Taxon effectively bridges the gap
between textual product descriptions and formal tax definitions. Our
multi-source data pipeline further ensures robustness to noisy supervision and
incomplete metadata, reflecting realistic e-commerce scenarios.

Comprehensive experiments on both internal and public datasets demonstrate that
Taxon substantially outperforms representative hierarchical and semantic
baselines across all evaluation levels. Notably, our analysis revealed that
residual errors primarily arise from structural inconsistencies in intermediate
nodes rather than from semantic misunderstanding at the leaf level. Motivated
by this insight, we introduced the \textit{RePath} procedure, which
reconstructs the hierarchical path from leaf predictions to enforce logical
path validity. This simple yet effective post-processing step consistently
enhances path-level $F_{1}$, confirming that accurate leaf predictions can
reliably recover the full taxonomic structure. Looking forward, we plan to
integrate \textit{RePath}-style structural correction into the training
objective, e.g., via differentiable path-validity constraints or joint
structured decoding, so that the model learns end-to-end hierarchical
consistency rather than relying on post-processing.


\section{Acknowledgments}
\label{sec:acknowledgments}
This work is supported by National Key R\&D Program of China under Grant No.
2024YFA1012700. It is also funded by the NSFC Project (No. 62306256) and the
Natural Science Foundation of Guangdong Province (No. 2025A1515010261).

\bibliographystyle{IEEEtran}
\bibliography{tax-code}

\end{document}